\documentclass[10pt,twocolumn,letterpaper]{article}
\usepackage{cvpr}
\usepackage{amsmath,amssymb,booktabs,multirow,graphicx}
\usepackage{microtype,url,natbib}
\makeatletter
\renewcommand\LenToUnit[1]{\strip@pt\dimexpr#1\relax}
\makeatother

\definecolor{cvprblue}{rgb}{0.21,0.49,0.74}
\usepackage[pagebackref,breaklinks,colorlinks,allcolors=cvprblue]{hyperref}
\def\paperID{*****}
\def\confName{CVPR}
\def\confYear{2027}
\newcommand{\method}{\textsc{FedNBO}}

\newcommand{\pp}{\,\mathrm{pp}}
\title{After the Federation: Benchmarking Knowledge Portability\\to Heterogeneous Newcomers}
\author{Xinyuan Zhao\\
Sun Yat-sen University}
\hypersetup{pdfauthor={Xinyuan Zhao}}
\newcommand{\revisionFemGain}{16.62}

\begin{document}

\maketitle

\begin{abstract}
Once historical clients become unavailable, a federation must serve newcomers through the artifact it leaves behind. How much of its knowledge remains useful in an architecture chosen afterward? We introduce \method, a seven-dataset benchmark that separates knowledge supply, policy identification, and private-model compilation under disjoint data roles. Controlled studies expose inflated oracle gains, architecture-dependent transfer reversals, and a gap between teacher accuracy and private-model utility. Ordinary public-response distillation provides a competitive reference without public task labels or newcomer samples: gains over strengthened local training are $23.52$ and $\revisionFemGain$ percentage points on 300 CelebA identities and 300 FedEMNIST writers, respectively. In a complete-pipeline FedEMNIST comparison, this reference reaches 97.13\% accuracy with 12.4 seconds of compilation per architecture, versus 97.03\% and 67.5 seconds for a HeteroFedTTA reconstruction at the same epoch budget. Equal-count public-input interventions extend support sensitivity to trainable visual backbones. Together, these results establish an auditable benchmark and a strong reference for evaluating the utility, cost and reliability of post-federation knowledge transfer.
\end{abstract}

\section{Introduction}

Federations must serve clients that arrive after historical participants become unavailable. New hospitals, devices and organizations may expose different modalities and choose architectures absent from training; restarting federated learning (FL)~\citep{mcmahan2017fedavg} may be impossible. The relevant endpoint is therefore the utility of the frozen artifact in a model chosen later.

\begin{figure*}[t]
\centering
\includegraphics[width=\textwidth]{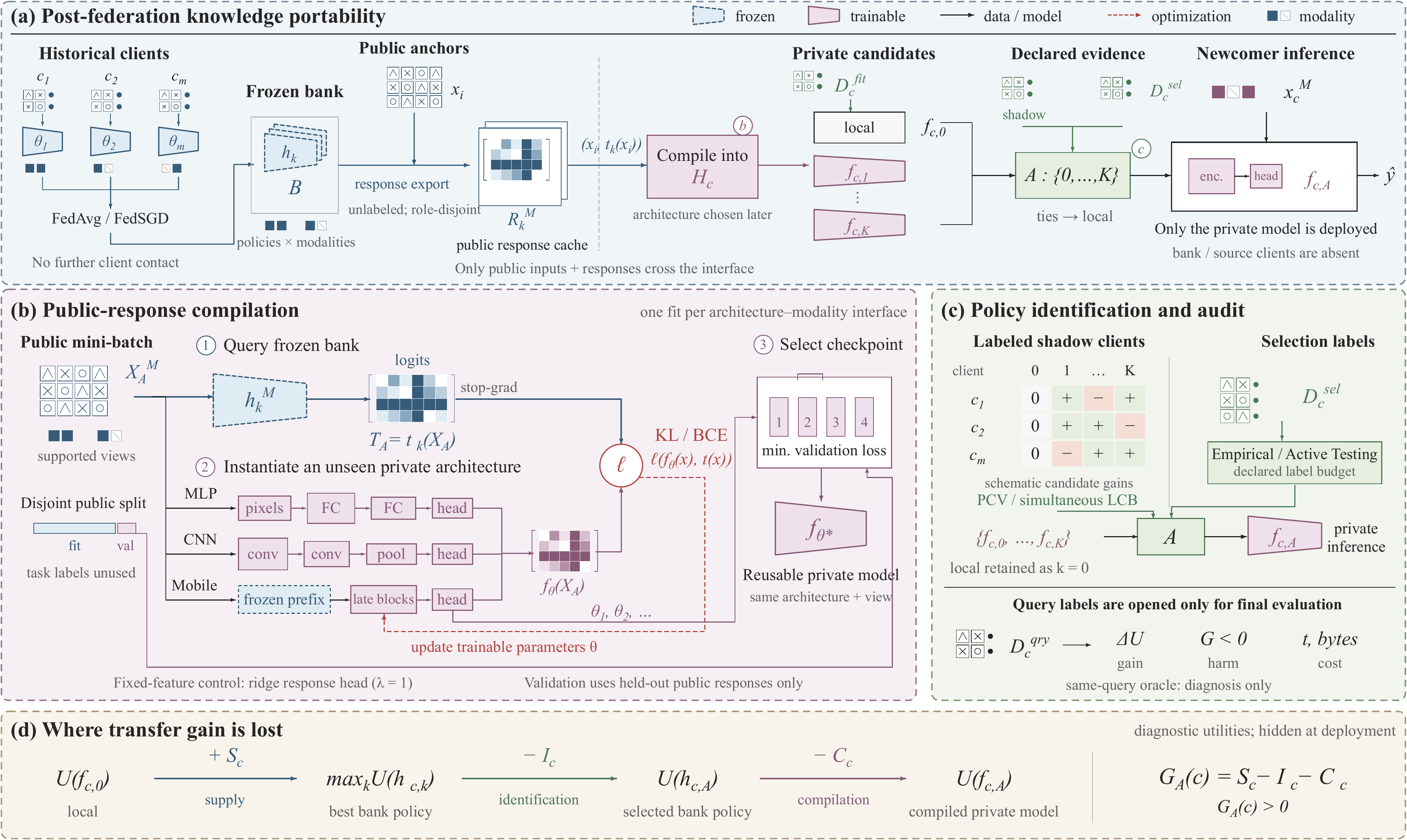}
\caption{\textbf{The \method{} protocol.} \textbf{(a)} A frozen bank exports public inputs/responses. \textbf{(b)} Private architectures learn from unlabeled responses; disjoint response validation selects checkpoints. Ridge is a separate control. \textbf{(c)} Shadow policies and newcomer selection-label methods are alternative routes; local wins ties and query labels are evaluation-only. \textbf{(d)} Gain separates supply, identification and compilation. Target utilities are hidden at deployment; glyphs are schematic.}
\label{fig:protocol}
\end{figure*}

Prior work handles late unlabeled clients, generates models in predefined spaces, or distills public responses into newcomer-chosen architectures~\citep{amosy2024odpfl,ye2024upfl,zakerinia2026flowdup,liu2025epfl}. We connect these capabilities to three deployment questions (Fig.~\ref{fig:protocol}): does useful knowledge exist, can allowed evidence identify it, and can the private model preserve it? Query-label reuse inflates opportunity; changing architecture reverses transfer on the same clients; sparse handoffs leave a 27.17-point teacher--private-model gap.

\method{} contributes: (1) a seven-dataset, role-disjoint benchmark with declared information access, unseen private architectures, accuracy--cost comparisons and client harm; (2) a supply--identification--compilation decomposition with corresponding identifiability boundaries; and (3) controlled evidence for a strong public-response reference. Ordinary KD gains $+23.52\pp$ on CelebA and $+\revisionFemGain\pp$ on FedEMNIST over strengthened local training. Complete-pipeline comparisons quantify its cost advantage, while equal-count interventions expose public-support sensitivity in trainable visual models. These results establish useful portability under explicit conditions rather than a universal transfer guarantee.

\section{Related Work}

\paragraph{Future clients and architectures.}
OD-PFL and EPFL generate joining-client models; UPFL/FedTTA meta-learns adaptation, and HeteroFedTTA distills public-data ensemble responses into newcomer-chosen models~\citep{amosy2024odpfl,liu2025epfl,ye2024upfl}. FLowDUP generates models in a trained parameter subspace~\citep{zakerinia2026flowdup}. pFedHN, PeFLL, HyperFLoRA and HyperFedZero use predefined parameter families~\citep{shamsian2021pfedhn,scott2024pefll,lu2024hyperfloRA,zhou2026hyperfedzero}; HAFL-GHN initializes withheld architectures from known layer types~\citep{litany2022ghn}. ATP adapts compatible models, and TFL studies out-of-federation generalization~\citep{bao2023atp,ma2024tfl}. Our contract overlaps HeteroFedTTA's base-response pathway; the contribution is its controlled deployment evaluation and failure decomposition.

\paragraph{Response interfaces and evidence.}
FedMD exchanges public responses among participating private models~\citep{li2019fedmd}; FedMBridge and FedAFD address heterogeneous multimodal participants~\citep{chen2024fedmbridge,tan2026fedafd}. We evaluate a frozen endpoint through complete pipelines and same-bank ablations, separating source effort, payload and newcomer cost. Active Testing supplies a generic label-efficient selector~\citep{kossen2021active}. Domain-adaptation theory motivates our distinction between response recovery and benefit under unconstrained target labels~\citep{ben2010theory}. The supplement maps each method's information and architecture requirements.

\section{Problem Formulation}

A completed federation exposes a frozen bank $B$; historical clients cannot be contacted or retrained. Newcomer $c$ has modality set $M_c$ and private hypothesis class $\mathcal H_c$, possibly unseen during federation training. Source and newcomer share label semantics and metric, but need not share representations or parameters. Fit labels $D_c^{\mathrm{fit}}$ train local model $f_{c,0}$. Bank policies $h_{c,1},\ldots,h_{c,K}$ are compiled into $f_{c,k}\in\mathcal H_c$; directly deployable policies have $f_{c,k}=h_{c,k}$.

A selector $A$ chooses $k\in\{0,\ldots,K\}$ using declared onboarding evidence, with $h_{c,0}=f_{c,0}$. Query labels $D_c^{\mathrm{qry}}$ are evaluation-only. For higher-is-better utility $U$, client gain is
\begin{equation}
 G_k(c)=U(f_{c,k};c)-U(f_{c,0};c)\in[-1,1].
\end{equation}
Negative transfer means $G_A(c)<0$. Fit and selection labels serve different roles: zero selection labels does not make a few-label-trained model zero-shot.

\paragraph{Public-response compilation.}
Identity-disjoint, unlabeled public anchors $A=\{x_i\}_{i=1}^n$ receive frozen logits $t(x_i)$ for each supported modality subset. Only $(x_i,t(x_i))$ crosses the interface. We fit fixed-feature ridge ($\lambda=1$) or ordinary neural response KD~\citep{hinton2015distilling}, including trainable late visual blocks; held-out responses select checkpoints without task labels or newcomer inputs. A compiled model is reusable across clients choosing the same architecture--modality interface. Historical models are absent at inference.

\paragraph{Three separable bottlenecks.}
Teacher-side supply is $S_c=\max_k U(h_{c,k};c)-U(f_{c,0};c)$. Identification loss is $I_c=\max_k U(h_{c,k};c)-U(h_{c,A};c)$, and compilation loss is $C_c=U(h_{c,A};c)-U(f_{c,A};c)$. Thus
\begin{equation}
 G_A(c)=S_c-I_c-C_c.\label{eq:accounting}
\end{equation}
Compilation addresses $C_c$, not the sign of $S_c-I_c$. A private-candidate oracle mixes these terms and is optimistic if query labels both select and score a policy; we also use label-disjoint selection/evaluation. All terms are diagnostic because target utility is hidden at deployment.

\section{Identifiability Boundary}

\textbf{Proposition 1 (unlabeled benefit is not identifiable).}
Let $f_0$ be local and $f_1,\ldots,f_K$ fixed candidates. A selector observes unlabeled target inputs, candidate predictions and their statistics. If it selects $f_j\ne f_0$ with positive probability where they disagree, then over unrestricted $P(Y\mid X)$ it cannot guarantee both nonnegative transfer and strictly positive gain. The same observations admit opposite label conditionals; a positive guarantee needs assumptions or label-linked evidence.

\textbf{Proposition 2 (anchor-span identifiability).}
Let $z(x)=[\phi_c(x);1]\in\mathbb R^d$ be a fixed representation with bias, and $Z_A$ its anchor matrix. If teacher scores satisfy $t(x)=w_*^\top z(x)$ on anchors and targets, anchor responses determine all target scores iff
\begin{equation}
 \operatorname{span}\{z(x):x\in\operatorname{supp}(P_c)\}
 \subseteq\operatorname{rowspan}(Z_A).\label{eq:span}
\end{equation}
Recovery error scales with the coefficient norm in $z(x)=Z_A^\top\alpha_x$; regularization also requires a representation-residual term. This is a fixed-feature response result, not a label-accuracy or neural-optimization guarantee. Exact recovery can copy a harmful teacher.

\textbf{Proposition 3 (shadow mean-gain certificate).}
Condition on a frozen bank and policies, $\mathcal F$. Suppose $m$ labeled shadow clients and a future client are conditionally independent draws from the same declared stratum. Policies are fixed before shadow labels are opened and $G_k(c)\in[-1,1]$. With probability at least $1-\delta$, simultaneously for $K$ candidates,
\begin{equation}
 \mathbb E[G_k(C)\mid\mathcal F]\geq\bar G_k-
 \sqrt{2\log(K/\delta)/m}.\label{eq:lcb}
\end{equation}
Hoeffding concentration and a union bound give the result~\citep{hoeffding1963probability}. A positive bound certifies expected future-client gain, not individual safety. For $m=3,K=5,\delta=.05$, the radius is $1.75$. Proofs and precise scope are supplementary.

\section{The \method{} Benchmark}

\subsection{Role-disjoint protocol}
Source clients train the bank; shadow (meta) clients select policies before held-out test scoring (Table~\ref{tab:datasets}). Prospective studies freeze plans before test access. Labeled revisions reuse evaluated cohorts but make new choices on shadow-development clients. The five-dataset policy audit repeats optimization with five seeds and linear/deep heads; UCI HAR uses linear models and temporal CNNs. Seeds are not independent clients.
\begin{table*}[t]
\centering
\small
\caption{Benchmark tasks.  TCGA-BRCA uses balanced accuracy; the others use accuracy.  UCF51-20 clients are group-preserving simulations; all other client units are natural.}
\label{tab:datasets}
\resizebox{\textwidth}{!}{%
\begin{tabular}{lcccp{3.0cm}p{5.7cm}}
\toprule
Dataset & classes & source/shadow/test & client unit & modalities & frozen bank $\rightarrow$ private models \\
\midrule
CREMA-D \citep{cao2014cremad} & 6 & 60/15/16 & actor (natural) & visual/audio features; some audio missing & global/cluster FedAvg MLPs $\rightarrow$ linear/three-layer MLP \\
MELD \citep{poria2019meld} & 4 & 15/3/3 & speaker (natural) & text/audio; dialogues remain disjoint & GloVe/audio MLPs $\rightarrow$ MobileBERT-feature linear/MLP \\
UCF51-20 \citep{soomro2012ucf101} & 51 & 12/4/4 & group silo (simulated) & video/audio features & FedAvg MLPs $\rightarrow$ linear/MLP \\
TCGA-BRCA \citep{tcga2018pancancer} & 2 & 7/3/3 & Tissue Source Site (natural) & WSI embedding/RNA expression & FedAvg MLPs $\rightarrow$ linear/MLP \\
UCI HAR \citep{anguita2013har} & 6 & 12/6/12 & subject (natural) & raw accelerometer/gyroscope windows & flattened-window MLPs $\rightarrow$ raw linear/temporal CNN \\
CelebA \citep{liu2015faceattributes} & 2 & 878/310/291; $+4{\times}300$ confirm & identity (natural) & raw RGB/facial landmarks & raw FedSGD ResNet-18 $\rightarrow$ MobileNetV3/EfficientNet-B0 \\
FedEMNIST \citep{cohen2017emnist,caldas2018leaf} & 10 & 1000/400/300 & writer (natural) & raw grayscale image & residual CNN $\rightarrow$ pixel MLP/small CNN \\
\bottomrule
\end{tabular}
}
\end{table*}
MELD assigns modalities by speaker and preserves dialogue separation. UCF51-20 preserves indivisible video groups; its clients are simulated. TCGA uses eligible Tissue Source Sites and class-stratified 60/20/20 splits. UCI HAR withholds gyroscope from half the subjects; FedEMNIST preserves writer identities and official within-writer partitions. Cohort and freeze receipts are supplementary.

\subsection{Candidates and baselines}
\textbf{Local} uses newcomer fit labels. \textbf{Joint KD} adds response KL with weights $0.1,1,10$; \textbf{pure KD} omits fit supervision. Candidates share private initialization within each cell. CREMA-D also tests global, clustered and uniform ensembles. Fit budgets are 32 labels on MELD, 16 on TCGA, two per class on HAR, and the existing private training partition on UCF51-20.

CelebA's label-blind hash prefix selects 4,096 images from 3,137 public identities outside evaluation clients. ResNet-18 responses train frozen-feature MobileNetV3-Small/EfficientNet-B0 ridge heads or late visual blocks; 512 held-out responses select neural checkpoints. Hard-label ridge and soft prototypes share anchors. The image pool is 28.2 MB; two float32 response tables total 64 KiB. FedEMNIST uses 4,096 images from 500 public writers to compile unseen pixel-MLP/CNN models. Neither compiler uses public task labels or newcomer inputs; the primary decentralized revision removes the historical public-label progression gate.

Selectors include fixed policies, empirical selection-label argmax, \textbf{Active Testing} (fixed ensemble surrogate, expected-loss acquisition, 10\% uniform floor and LURE correction~\citep{kossen2021active}), \textbf{PCV} (one shadow-selected policy, requiring nonnegative mean gain for every shadow client), and Eq.~\ref{eq:lcb}. Local wins ties; same-query oracles are diagnostic. HAR additionally executes PeFLL-compatible generation from 12 labeled descriptors and ATP-compatible one-step entropy adaptation from 20 unlabeled inputs~\citep{scott2024pefll,bao2023atp}; their supported interfaces and references differ.

\subsection{Metrics and statistics}
Gain is in percentage points (pp). A cell is a client--architecture--seed combination. Historical client harm averages architectures/seeds first; complete-pipeline harm is architecture-specific after seed averaging. Reported 95\% percentile bootstrap intervals resample clients, conditional on trained models. Reused-cohort revisions, post-hoc controls and prospective confirmations retain their distinct evidence status.
\section{Results}

\subsection{Useful transfer across tasks and domains}
\begin{table*}[t]
\centering\small
\caption{\textbf{Public-response utility and cost.} \textbf{A:} strengthened local controls on 300 clients per dataset (reused-cohort revisions). CelebA matches late blocks and eight fit labels; FedEMNIST uses 20 labels/1,440 local updates. \textbf{B:} separate 100-epoch, three-seed pipeline comparison; harm is MLP/CNN after seed averaging. Source costs differ; the full comparator reconstructs HeteroFedTTA.}
\label{tab:utility-cost}\label{tab:primary-revision}
\begin{minipage}[t]{.49\textwidth}\centering
\textbf{A. Benefit over strengthened local training}\\[3pt]
\begin{tabular}{lrrr}\toprule Dataset & local & compiled & gain [95\% CI] \\ \midrule
CelebA & 62.94 & 86.46 & 23.52 [21.69,25.38] \\
FedEMNIST & 80.27 & 96.89 & 16.62 [15.35,17.86] \\
\bottomrule\end{tabular}
\end{minipage}\hfill
\begin{minipage}[t]{.49\textwidth}\centering\setlength{\tabcolsep}{3pt}
\textbf{B. Complete pipeline: accuracy, harm and cost}\\[3pt]
\begin{tabular}{lrrr}\toprule Pipeline & acc. (\%) & harm MLP/CNN & time (s) \\ \midrule
Original bank + KD & 97.13 & 12.0/2.0 & 12.4 \\
Meta bank + KD & 97.09 & 12.7/2.3 & 11.9 \\
Meta bank + dual & 97.03 & 12.3/2.7 & 67.5 \\
\bottomrule\end{tabular}
\end{minipage}
\vspace{3pt}\parbox{.98\textwidth}{\footnotesize Accuracy is \%, gain is pp. Panel A FedEMNIST compilers use 50 epochs; B uses 100, with 200-epoch sensitivity in Table~\ref{tab:budget-main}. Compilation time is per architecture on one RTX 4090. Intervals in A resample clients after averaging architectures and FedEMNIST seeds.}
\end{table*}

\paragraph{Strengthened controls.}
CelebA uses a client-uniform FedSGD ResNet-18 bank trained on 878 identities; label-independent identity hashing assigns landmark availability. Private MobileNetV3/EfficientNet architectures are absent from federation training. Independent 300-identity confirmations find $+16.63\pp$ for ridge and 86.46\% for trainable blocks, the latter adding $5.19\pp$ over frozen ridge. The primary revision matches local ImageNet initialization and trainable blocks, selecting its optimization on shadow identities; compilation retains $+23.52\pp$ (Table~\ref{tab:utility-cost}A). Even a query-selected local ridge/finetuning oracle leaves $+18.52\pp$; this is diagnostic only.

FedEMNIST's primary revision uses three client-local FedSGD teachers with GroupNorm and response-only compilation. On 300 writers it reaches 96.89\% versus 80.27\% for 20-label local training with shadow-selected learning rates and 1,440 updates. Seed standard deviation is $0.16\pp$; three models do not make 900 independent writers. Architecture-averaged harm is 1.7--3.0\%. The centralized-source audit remains a historical reference.

\begin{figure*}[t]
\centering
\includegraphics[width=\textwidth]{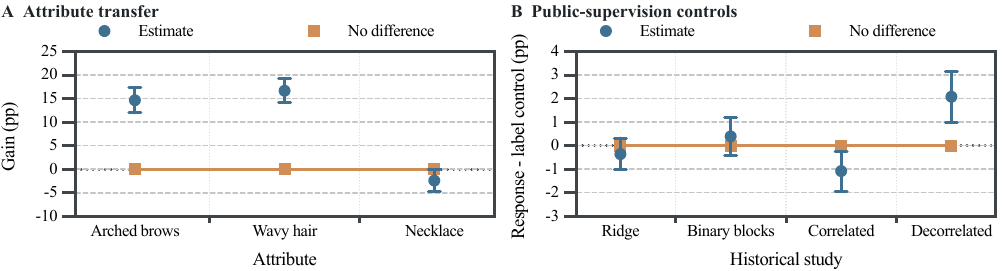}
\caption{\textbf{Task and supervision controls.} \textbf{A:} Attribute gains on 300 decorrelated CelebA identities include the negative Necklace result. \textbf{B:} Response-minus-public-label-control gains in separate historical studies. These post-confirmation supervised controls are not compiler inputs. Bars: 95\% client-bootstrap intervals; cohorts and local references differ across conditions.}
\label{fig:task-controls}
\end{figure*}

\paragraph{Task and supervision controls.}
A correlated-attribute confirmation gains $19.53\pp$ with 7.7\% harm. A later, disjoint 300-identity study selects attributes with $|\mathrm{corr}(Y,\mathrm{Smiling})|\leq.10$ and pairwise correlation $\leq.20$ using metadata: mean gain is $9.66\pp$ [8.09,11.20], but harm is 19.7\%. Figure~\ref{fig:task-controls}A retains the negative Necklace result; these cohorts do not identify a paired decorrelation effect. Separately supervised public-label controls (Fig.~\ref{fig:task-controls}B) yield response-minus-control differences of $-0.35,+0.40,-1.08,+2.09\pp$ across ridge, binary, correlated and decorrelated studies. Responses are competitive, not uniformly superior to public labels.
\subsection{Honest policy identification}
\begin{figure*}[t]
\centering
\includegraphics[width=\textwidth]{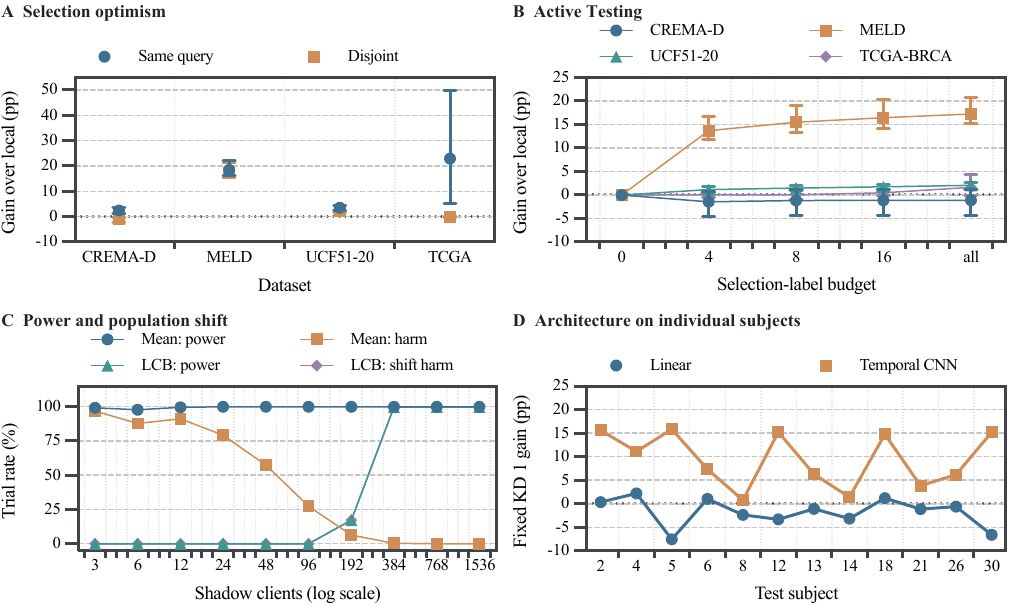}
\caption{\textbf{Four onboarding audits.} \textbf{A:} Same-query versus label-disjoint selection (95\% newcomer intervals). \textbf{B:} Active Testing across label budgets. \textbf{C:} Mean selection and LCB power/harm under declared or shifted populations (50,000 trials). \textbf{D:} Fixed KD-1 gains on 12 HAR subjects, averaged over seeds within each architecture.}
\label{fig:evidence-atlas}
\end{figure*}
\begin{table*}[t]
\centering\small
\caption{\textbf{Fixed policies and shadow-selected deployment.} Gain/CI: pp; harm: \%. Policies 1--4 are global/cluster 0/cluster 1/uniform on CREMA-D and KD 0.1/1/10/pure KD elsewhere. PCV uses only shadow labels; all LCB rules abstain. Test cohorts: MELD/TCGA, 3 clients; UCF, 4 simulated silos.}
\label{tab:selection-main}
\setlength{\tabcolsep}{5pt}
\begin{tabular}{@{}lrrrrllr@{}}
\toprule
& \multicolumn{4}{c}{Fixed-policy gain} & \multicolumn{3}{c}{PCV deployment} \\
\cmidrule(lr){2-5}\cmidrule(l){6-8}
Dataset & 1 & 2 & 3 & 4 & Choice & Gain [95\% CI] & Harm \\
\midrule
CREMA-D & $-7.08$ & $-6.78$ & $-8.80$ & $-6.87$ & local & $0.00$ [0.00,0.00] & 0.0 \\
MELD & $+11.16$ & $+16.05$ & $+17.66$ & $+17.89$ & pure KD & $+17.89$ [15.30,22.03] & 0.0 \\
UCF51-20 & $+1.90$ & $+1.57$ & $-1.18$ & $-1.45$ & local & $0.00$ [0.00,0.00] & 0.0 \\
TCGA-BRCA & $+13.33$ & $+17.62$ & $+17.18$ & $+14.21$ & joint KD 0.1 & $+13.33$ [$-1.31$,30.00] & 33.3 \\
UCI HAR & $+3.23$ & $+3.88$ & $+1.98$ & $+1.69$ & local & $0.00$ [0.00,0.00] & 0.0 \\
\bottomrule
\end{tabular}
\end{table*}

Same-query versus label-disjoint gain falls from $+2.49$ to $-1.01\pp$ on CREMA-D and $+22.98$ to $-0.08\pp$ on TCGA (Fig.~\ref{fig:evidence-atlas}A); UCF51-20 retains $+2.12\pp$ from $+3.56\pp$. Candidate existence and policy identification differ: all fixed CREMA-D policies are harmful, all MELD policies help, and HAR retains positive fixed gains despite PCV abstention (Table~\ref{tab:selection-main}). TCGA has a useful fixed KD-1 policy ($+17.62\pp$), but small, imbalanced selection cohorts fail to identify it per client.

\paragraph{More labels help selectively.}
Four Active Testing labels recover $13.68\pp$ on MELD and $1.14\pp$ on UCF51-20; additional labels improve both (Fig.~\ref{fig:evidence-atlas}B). CREMA-D stays negative with all selection labels, while TCGA reaches only $1.67\pp$. Labels cannot create transferable candidates or make a shifted cohort representative.

\paragraph{Shadow evidence and shift.}
PCV abstains on CREMA-D, UCF51-20 and HAR. On TCGA it selects KD-0.1 and gains $13.33\pp$, yet harms site A2 and has a client interval crossing zero. Three nonnegative shadow sites did not protect every future site. The formal LCB abstains throughout: a true gain of .20 requires $m\geq231$ for a positive bound at the mean. In 50,000 bounded-gain trials, deployment reaches 99.8\% at $m=384$ under future mean .20; shifting that mean by $-.25$ causes harmful deployment in 99.8\% of trials (Fig.~\ref{fig:evidence-atlas}C). This is a population-assumption audit, not another benchmark.

\subsection{Private architecture and handoff loss}
\begin{table*}[t]
\centering\small
\caption{\textbf{HAR architectures, selection labels and compatible baselines.} \textbf{A:} sealed 12-subject audit; gain/CI: pp, harm: \%. Oracle is diagnostic. \textbf{B:} separate supported slices with different references; PeFLL uses 12 labeled descriptors, ATP an unlabeled batch of 20. Accuracy: \%.}
\label{tab:architecture-main}
\textbf{A. Fixed transfer and disagreement-label audit}\par\smallskip
\setlength{\tabcolsep}{7pt}
\begin{tabular}{@{}lrrrr@{}}
\toprule
Policy & Gain [95\% CI] & Harm & Linear gain & CNN gain \\
\midrule
Fixed joint KD 1 & $+3.88$ [2.19,5.51] & 16.7 & $-1.73$ & $+9.49$ \\
Disagreement audit, 6 labels & $+3.22$ [1.83,4.64] & 8.3 & $-1.08$ & $+7.52$ \\
Disagreement audit, 24 labels & $+4.47$ [2.71,6.34] & 8.3 & $-0.93$ & $+9.87$ \\
Same-query oracle (diagnostic) & $+8.03$ & 0.0 & $+2.89$ & $+13.18$ \\
\bottomrule
\end{tabular}\par\medskip
\textbf{B. Named methods on supported interfaces}\par\smallskip
\setlength{\tabcolsep}{6pt}
\begin{tabular}{@{}llrrrr@{}}
\toprule
Method & Reference & Ref. acc. & Method acc. & Gain [95\% CI] & Harm \\
\midrule
PeFLL & FedAvg + fit (12 labels) & 44.4 & 51.2 & $+6.79$ [3.35,9.95] & 16.7 \\
ATP & Compatible global model & 46.02 & 45.97 & $-0.05$ [$-0.69$,0.59] & 50.0 \\
\bottomrule
\end{tabular}
\end{table*}

HAR freezes roles, missingness, five policies and budgets before opening 12 test subjects; a shadow-only amendment sets two fit labels per class. Fixed KD-1 gains $3.88\pp$ overall but loses $1.73\pp$ in linear models and gains $9.49\pp$ in temporal CNNs (Fig.~\ref{fig:evidence-atlas}D). The paired interaction is $11.21\pp$ [7.57,15.17], $p=.0005$ by exact sign-flip. Six disagreement labels halve harm from 16.7\% to 8.3\%; 24 retain that harm and raise gain to $4.47\pp$ (Table~\ref{tab:architecture-main}A). Neither removes negative linear-stratum gain. Architecture routing discovered here remains retrospective.

Compatible PeFLL and ATP slices have different local references (Table~\ref{tab:architecture-main}B): PeFLL gains $6.79\pp$ but harms 16.7\%; ATP changes mean accuracy little ($-0.05\pp$) while harming 50.0\%. They are not a common-baseline ranking. On 291 sealed CelebA identities, sparse prototype handoff reaches 65.23\% versus 63.49\% local and 92.40\% teacher, leaving a $27.17\pp$ [25.17,29.17] portability gap and 37.5\% harm. A six-transformation control does not resolve it ($-0.47\pp$ [$-2.19$,1.09]).

\subsection{Public-input quantity, support and compression}
\begin{figure*}[t]
\centering
\includegraphics[width=\textwidth]{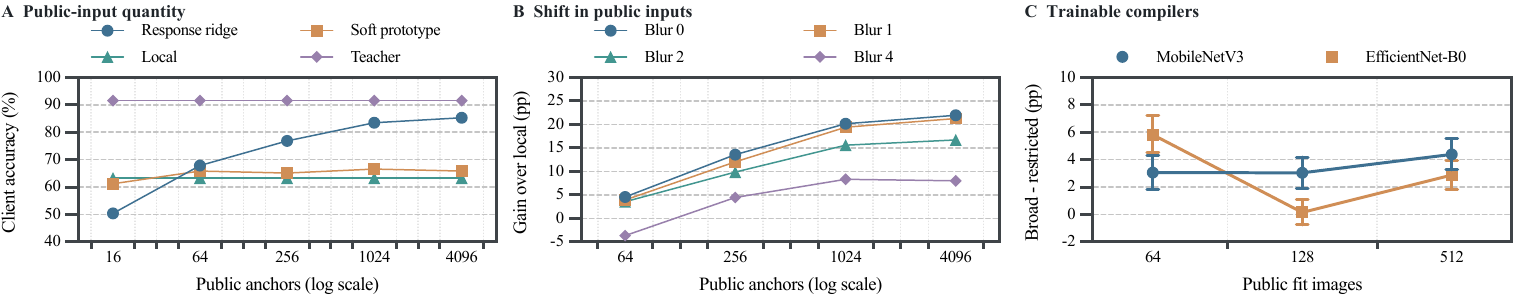}
\caption{\textbf{Public inputs and private transfer.} On the same 270 shadow identities: \textbf{A:} anchor-count frontier; \textbf{B:} blur applied only to public inputs; \textbf{C:} broad-minus-restricted neural support with equal counts, 336 updates and 512 extra validation responses. Bars in C are paired 95\% client intervals conditional on three draws; EfficientNet at 128 is inconclusive.}
\label{fig:anchor-portability}
\end{figure*}
\begin{table*}[t]
\centering\small
\caption{\textbf{Equal-count public-support ablation.} Broad/restricted pools share teacher, initialization, 336 updates, 512 validation responses and 270 identities. Accuracy/harm: \%; differences: pp. CIs resample clients after averaging three draws. $\Delta$KL is broad minus restricted; harm is broad/restricted.}
\label{tab:support-main}
\setlength{\tabcolsep}{5pt}
\begin{tabular}{@{}lrrrrrr@{}}
\toprule
Private backbone & Fit inputs & Broad acc. & Restr. acc. & Difference [95\% CI] & $\Delta$KL & Harm \\
\midrule
MobileNet blocks & 64 & 72.07 & 69.01 & 3.06 [1.84,4.32] & $-0.0490$ & 27.4/32.2 \\
& 128 & 78.46 & 75.42 & 3.04 [1.93,4.20] & $-0.0426$ & 17.8/24.8 \\
& 512 & 85.32 & 80.94 & 4.38 [3.26,5.54] & $-0.0861$ & 10.4/13.7 \\
\addlinespace[3pt]
EfficientNet blocks & 64 & 75.96 & 70.12 & 5.83 [4.51,7.21] & $-0.0242$ & 25.6/35.2 \\
& 128 & 82.19 & 82.04 & 0.15 [$-0.76$,1.10] & $+0.0248$ & 14.1/14.4 \\
& 512 & 87.82 & 84.94 & 2.89 [1.84,3.94] & $-0.0667$ & 8.5/13.0 \\
\bottomrule
\end{tabular}
\end{table*}

On 270 shadow identities, ridge accuracy rises from 50.44\% at 16 public anchors to 85.28\% at 4,096, versus 63.31\% local and 91.57\% teacher (Fig.~\ref{fig:anchor-portability}A). Controlling budget, backbone and modality removes the coverage--gain association ($\rho=-.012$ [$-.095$,.070]). Blurring only public inputs reduces 4,096-anchor gain from $21.97$ to $8.01\pp$ at radius 4 and raises harm from 5.6\% to 24.8\% (Fig.~\ref{fig:anchor-portability}B).

\paragraph{Support beyond count.}
An equal-count, fixed-feature intervention compares broad inputs with the upper quartile of pose asymmetry. Broad support adds $4.61\pp$ [3.67,5.56] at 64 anchors and $3.90\pp$ [2.92,4.90] at 128 across three draws. Holding initialization, 336 neural updates and 512 validation responses fixed extends this comparison to trainable blocks (Table~\ref{tab:support-main}; Fig.~\ref{fig:anchor-portability}C). At 512 inputs, harm falls from 13.7/13.0\% to 10.4/8.5\% for MobileNet/EfficientNet. EfficientNet at 128 remains inconclusive despite worse response KL. These neural findings are empirical, not consequences of Proposition~2.

\paragraph{Matched and post-hoc controls.}
With identical initial FP32 features, inputs and $\lambda=1$, late-block training exceeds closed-form ridge by $5.52,6.51,6.47\pp$ at 64/128/512 inputs. On separate cohorts, hard-label/response ridge reach 80.25/80.85\%, while soft prototypes reach 65.83\%. Neither pivoted QR nor target-nearest anchor selection consistently beats a hash prefix. A shadow-selected one-bit quantizer gives validation accuracy 84.91\% versus float32 85.28\%; on a reused 300-identity cohort it gives 85.71\%, $+22.56\pp$ and 7.0\% harm with a 1,024-byte response digest. The 28.2-MB image pool remains required; this post-hoc compression audit implies no privacy guarantee. Full control grids are supplementary.

\subsection{Complete-pipeline cost and client harm}
\begin{table*}[t]
\centering\small
\caption{\textbf{FedEMNIST compiler-budget sensitivity.} Entries are 100/200 epochs with matched inputs, initialization, optimizer and batch order. Accuracy/harm: \%; harm averages three seeds by architecture. Meta + dual reconstructs HeteroFedTTA. Timings exclude I/O/loading; source effort differs and is reported separately.}
\label{tab:budget-main}\label{tab:complete-pipeline}
\setlength{\tabcolsep}{7pt}
\begin{tabular}{@{}llrrrr@{}}
\toprule
Architecture & Pipeline & Accuracy & Harm & Compile (s) & Target (ms) \\
\midrule
MLP & Original + KD & 95.87/95.82 & 12.0/12.0 & 11.8/23.1 & 0.71/0.66 \\
& Meta + KD & 95.74/95.97 & 12.7/11.0 & 11.0/22.4 & 0.66/0.66 \\
& Meta + dual & 95.76/95.82 & 12.3/12.0 & 67.1/132.2 & 6.11/6.23 \\
\addlinespace[3pt]
CNN & Original + KD & 98.39/98.33 & 2.0/2.3 & 12.9/25.7 & 0.79/0.77 \\
& Meta + KD & 98.44/98.49 & 2.3/2.7 & 12.8/25.3 & 0.69/0.69 \\
& Meta + dual & 98.29/98.42 & 2.7/2.0 & 67.9/136.5 & 42.88/10.94 \\
\bottomrule
\end{tabular}
\end{table*}

At 100 epochs, original-bank KD reaches 97.13\% with 12.4 s compilation per architecture versus 97.03\% and 67.5 s for the HeteroFedTTA reconstruction (Table~\ref{tab:utility-cost}B). Full-minus-minimal gain is $-0.11\pp$ [$-0.28$,0.07]; at 200 epochs it is $0.04\pp$ [$-0.12$,0.21], not evidence of equivalence. Table~\ref{tab:budget-main} retains both architectures and budgets: doubling optimization roughly doubles compilation without uniform accuracy gain. Same-bank KD separates source improvement from target adaptation; at 200 epochs its advantage over original-bank KD is $0.15\pp$ [0.04,0.27].

\paragraph{Amortization and source effort.}
For $J$ interfaces and $N$ arrivals, stage costs satisfy
\begin{equation}
 C_{\mathrm{total}}=C_{\mathrm{bank}}+J\bar C_{\mathrm{compile}}+N\bar C_{\mathrm{target}}.\label{eq:deployment-cost}
\end{equation}
Bank cost includes export; communication is separate. Original source training takes 85.7--91.6 s, selected meta checkpoints add 2,656.9--10,668.1 s, and source-budget search costs 12,463.9--43,954.6 s. Uncached minimal/full payloads are 3.22/3.38 MiB (responses: 160/320 KiB). These are matched onboarding budgets, not equal source effort or distributed latency; one reconstructed source run reaches its cap before plateau. Reuse determines the tradeoff.

\paragraph{Who is harmed?}
Original-bank KD harms 12.0\% of MLP writers versus 2.0\% of CNN writers. In the paired CelebA historical-ridge audit, updating blocks reduces compiler-induced harm from 6.3\% to 2.0\% and compiler loss from $6.90$ to $1.71\pp$; inherited harm stays near 4\%. Image-only and image--landmark gains remain $24.74/22.03\pp$. The decorrelated task retains 19.7\% harm (10.3\% inherited, 9.3\% compilation): Necklace's $1.19\pp$ teacher supply cannot absorb $3.60\pp$ compilation loss. Query-set rates are not individual-risk guarantees.

A diagnostic CREMA-D public-actor audit instead has insufficient supply: local/teacher/64-anchor accuracy is 80.68/55.19/53.13\%. Broad-minus-restricted gains at 64/96 inputs cross zero, and both pools harm all ten validation actors. Only three have audio. Better response recovery cannot rescue a harmful teacher. Primary mean evidence comes from 270--300-client image cohorts; HAR and the remaining tasks expose mechanism and selection boundaries.

\section{Scope and Implications}

Report teacher utility against a credible local baseline, separate selection from scoring, and test the intended private architecture at a declared budget. CelebA covers image-only and image-derived landmarks; FedEMNIST tests unimodal portability. General independent-sensor multimodal transfer remains distinct. The \texttt{fednbo.py} entry point binds cohort identities, protocols, result hashes and reproduction commands.

\section{Limitations}

MELD/TCGA have three test clients, UCF51-20 clients are simulated, and four tasks use frozen features. The image studies use in-domain anchors; cross-domain support, response privacy and prospective individual safety remain open. Proposition~2 concerns fixed linear representations. The HeteroFedTTA reconstruction has finite source budgets; revisions reuse evaluated cohorts. The supplement preserves superseded centralized FedEMNIST and public-label-gate protocols alongside the client-local replications.

\section{Conclusion}

\method{} makes post-federation portability measurable through knowledge supply, policy identification and private-model compilation. Controlled experiments reveal where apparent opportunities disappear and where a simple public-response handoff succeeds. Its strong image-task results, competitive complete-pipeline accuracy and lower compilation cost establish a useful reference for future methods. Public-support interventions and client-level harm identify conditions those methods must confront to improve deployment outcomes.\label{end:main-body}

{
\small
\bibliographystyle{ieeenat_fullname}
\bibliography{references}
}
\end{document}


\onecolumn
\begin{center}
{\Large\bfseries Supplementary Material: After the Federation}\par
\end{center}
\vspace{0.5em}
\renewcommand{\thetable}{S\arabic{table}}
\renewcommand{\thefigure}{S\arabic{figure}}

\section{Protocol, Cohorts and Evidence Status}
\label{sec:protocol-details}

This supplement follows the four experimental questions in the main paper: selection, private-model realization, public-input support, and cost/benefit. The main paper presents the principal policy, architecture, support, and budget contrasts. The consolidated tables here retain their original context and additional shadow, historical-control, and source-search rows. Full grids and individual seeds remain in the accompanying JSON/CSV files.

\paragraph{Roles and uncertainty.}
Source clients train the bank; shadow-development clients choose recipes; shadow-validation and test clients evaluate frozen decisions. Revisions explicitly reuse previously evaluated cohorts. The primary revised CelebA and FedEMNIST results are not additional prospective confirmations. Confidence intervals resample natural clients after averaging architectures and seeds, unless architecture is a displayed stratum. They quantify client sampling conditional on the trained models. Three teacher seeds do not turn 300 writers into 900 clients. MELD and TCGA have only three test clients each; UCF51-20 uses four simulated test silos. These studies diagnose boundary cases rather than establish population-wide success.

\paragraph{Natural-client protocols.}
The original CelebA protocol has 1,479 eligible identities: 878 source, 310 shadow (40 development, 270 validation) and 291 test. Its metadata-only v2 amendment increased the minimum class count from eight to ten within already frozen roles before test-pixel access. Each query has four examples per class; selection has two. Local fitting uses four labels per class. Landmark availability is assigned by an identity hash independently of labels. Subsequent ridge, trainable-block, correlated-attribute and decorrelated-attribute confirmations each use 300 new identities, disjoint from all preceding cohorts and public identities. Their image counts are 6,881, 6,959, 6,468 and 8,016. The current primary baseline revision uses the trainable-block cohort. The registry verifies exact identities and result cells for all five cohorts.

FedEMNIST uses the digits-only TensorFlow Federated release. An eligible writer has two training examples per digit and at least seven represented digits in its official test split. A metadata hash assigns 1,000 source, 500 public, 100 development, 300 validation and 300 test writers from 2,582 eligible writers. The 20 local labels are fixed training examples; official per-writer test images are query-only. Public-image ordering excludes task labels. UCF groups remain indivisible across data roles; the earlier clip-split UCF20 result is excluded from the benchmark.

\paragraph{Frozen chronology.}
Table~\ref{tab:confirmation-timeline} binds each original confirmation policy to the first complete result artifact. The guarded runner records the consumed policy hash; the verifier checks policy, receipt and result agreement. Local filesystem times (UTC+8) are repository records, not independent trusted timestamps. The original FedEMNIST progression gate used public labels and a centralized teacher; the later primary replications remove both dependencies. Task amendments, historical results and strengthened-baseline revisions remain distinguishable in the release.

\begin{table}[htbp]\centering\small
\caption{Original confirmation chronology. Hash prefixes identify full policy hashes in the receipts. Revised baselines reuse these cohorts rather than create new confirmations.}
\label{tab:confirmation-timeline}
\resizebox{.99\textwidth}{!}{\begin{tabular}{lllll}
\toprule
Study & roles & frozen choice & policy/result time & policy SHA \\
\midrule
CelebA sparse handoff & 878/310/291 identities & fit pseudo-label prototype; 8 labels & 09-26 23:53/09-26 23:55 & \texttt{9f2392c1} \\
CelebA ridge & new 300 identities & 4,096 anchors; ridge alpha 1 & 09-27 01:26/09-27 01:27 & \texttt{59f616ab} \\
CelebA trainable blocks & second new 300 & late blocks; 8 epochs; response selection & 09-27 02:07/09-27 02:09 & \texttt{561559fa} \\
CelebA 3-label & third new 300 & 3 incidence labels; late blocks & 09-27 10:39/09-27 10:40 & \texttt{a0b9c7e6} \\
CelebA decorrelated 3-label & fourth new 300 & correlation rule; inherited blocks & 09-27 12:24/09-27 12:26 & \texttt{9a2de6ac} \\
FedEMNIST & 1000/400/300 writers & 4,096 anchors; 2 architectures; 20-label local & 09-27 13:53/09-27 14:02 & \texttt{6f66f2a4} \\
\bottomrule
\end{tabular}
}
\end{table}

\paragraph{Initial policy-audit optimization.}
CREMA-D, MELD, UCF51-20 and TCGA use seeds 4201/4301/4401/4501/4601, linear and three-layer private heads, and respectively 160/200/200/240 updates. AdamW rates are $3\times10^{-3}$ for linear and $10^{-3}$ for deep heads; hidden width is 64 with normalization and GELU. TCGA uses balanced loss and accuracy. UCI HAR uses seeds 6101/6201/6301/6401/6501, 40 FedAvg rounds, flattened-window teachers and linear/temporal-CNN private models trained for 150 steps on two labels per class. The natural client, rather than the optimization cell, is the resampling unit.

\section{Useful Transfer and the Historical Controls}
\label{sec:confirmation-details}

Main Table~2 contains the strengthened primary baselines and the complete-pipeline comparison. Table~\ref{tab:cohorts} consolidates the earlier confirmation endpoints, so that the baseline changes and independent cohorts can be inspected without six separate result tables. These are paired gains within rows; improvements across different rows cannot be attributed to a single intervention.

\begin{table}[htbp]\centering\small
\caption{Historical cohort endpoints, distinct from the strengthened references in main Table~2. Each row retains its own local baseline; rows are not paired interventions. Accuracy is \%, gain is pp; the two attribute studies use balanced accuracy. Harm averages architectures; decorrelated ties use the recorded correction. The original FedEMNIST teacher was centralized.}
\label{tab:cohorts}
\setlength{\tabcolsep}{4pt}\begin{tabular}{lrrrrr}
\toprule
Study (clients) & Local & Private & Teacher & Gain [95\% CI] & Harm (\%) \\ \midrule
CelebA sparse (291) & 63.49 & 65.23 & 92.40 & 1.74 [0.15,3.22] & 37.5 \\
CelebA ridge (300) & 64.23 & 80.85 & 88.13 & 16.63 [14.63,18.63] & 15.3 \\
CelebA blocks (300) & 63.15 & 86.46 & 88.17 & 23.31 [21.33,25.27] & 6.0 \\
CelebA correlated (300) & 55.95 & 75.48 & 76.44 & 19.53 [17.92,21.19] & 7.7 \\
CelebA decorrelated (300) & 56.40 & 66.07 & 67.19 & 9.66 [8.09,11.20] & 19.7 \\
FedEMNIST original (300) & 77.90 & 96.75 & 99.34 & 18.85 [17.49,20.20] & 1.3 \\
\bottomrule\end{tabular}

\end{table}

\paragraph{Historical and private visual models.}
A client-uniform FedSGD ResNet-18 starts from ImageNet weights and updates layer 4 plus its image/landmark head; BatchNorm statistics remain frozen. Its selected source-query accuracy is 91.51\%. The original sparse handoff forms prototypes from hard teacher labels on local fit images, overwriting the eight true labeled examples. It never calls the teacher on query inputs. On the 270 validation identities it gains $+2.41$ points, while direct teacher accuracy is 91.57\%; its sealed 291-identity endpoint appears in Table~\ref{tab:cohorts}. A six-transformation development control changes gain by $-0.47\pp$ [$-2.19$,1.09], leaving the sparse-handoff limitation unresolved.

Public ridge fits historical log-odds with fixed $\lambda=1$ on frozen MobileNetV3-Small or EfficientNet-B0 features. The neural compiler updates MobileNet blocks 9--12 or EfficientNet blocks 6--8 plus their landmark-aware heads, leaving earlier ImageNet blocks fixed. Each public image supplies image-only and image--landmark views. BCE matches historical probabilities; 3,584 images fit and 512 response-only images select checkpoints. The eight-epoch visual-block recipe and checkpoints were fixed before the second confirmation; no public task label or newcomer sample enters compilation. Compilers are shared within an architecture/modality interface.

\paragraph{Strengthened local controls.}
The current CelebA local reference has the identical ImageNet initialization and trainable blocks, a ridge-initialized head and eight labels. Forty development identities select learning rates $10^{-4}/10^{-3}$, 10/30/100 updates and horizontal flips with consistent landmark remapping. A query-label-privileged local oracle choosing between ridge and finetuning still leaves $+18.52\pp$ [16.69,20.38] for compilation; it is a diagnostic, not a deployment rule. Three of 600 original ridge cells change near-zero predictions between half- and single-precision features. The revision keeps the archived compiler endpoint fixed and records this numerical amendment separately.

The primary FedEMNIST teachers use three independent seeds, 1,200 FedSGD rounds, eight writers per round and 64 examples per writer. Each forward/backward pass accesses one writer; the server averages gradients and uses AdamW. GroupNorm avoids cross-writer BatchNorm statistics; source-query accuracy selects the teacher. Fifty-epoch private KD uses AdamW at $8\times10^{-4}$ and response-only validation. A 100-writer development grid first tested 120/360/720 local steps and rates $5\times10^{-4}/2\times10^{-3}$. A logged upper-bound amendment added 1,440 steps before the new local test summary was inspected. MLP/CNN select 1,440 steps at these two respective rates. The three compiled accuracies are 96.92/97.03/96.71\%, with gains 16.65/16.76/16.44 points; mean gain is 16.62 with seed SD 0.16 points. Individual-run harm and model records remain in the numeric export.

\paragraph{Task and supervision controls.}
The incidence-selected correlated attributes are High Cheekbones, Mouth Slightly Open and Oval Face; source correlations with Smiling are .657/.572/.117. The later task requires $|\mathrm{corr}|\leq.10$ with Smiling and pairwise $|\mathrm{corr}|\leq.20$, selecting Arched Eyebrows, Wavy Hair and Wearing Necklace from incidence metadata. Their gains are $+14.70\pp$ [11.99,17.38], $+16.69\pp$ [14.18,19.24], and $-2.40\pp$ [$-4.67$,$-0.10$]. Local references use at most 12 labeled images and development-selected nearest centroids. The corresponding 20/75 and 20/91 development/validation cohorts never join confirmation roles.

Public-label oracle controls are separate supervised experiments, performed after the corresponding CelebA confirmations. For response ridge, binary blocks, correlated blocks and decorrelated blocks, oracle accuracies are 81.21/86.06/76.56/63.98\%; response-minus-oracle differences are $-0.35\pp$ [$-1.00$,0.33], $+0.40\pp$ [$-0.40$,1.19], $-1.08\pp$ [$-1.92$,$-0.24$], and $+2.09\pp$ [0.98,3.15]. The original FedEMNIST supervised public model reaches 95.54\%; response KD exceeds it by 1.21 points [0.82,1.61]. These controls constrain the role of public supervision, without changing the information available to the response method.

\section{Can Allowed Evidence Select a Useful Policy?}
\label{sec:selection-details}

\begin{table}[htbp]\centering\small
\caption{Fixed-policy opportunity and deployable population selection. The four gains are global/cluster~0/cluster~1/uniform for CREMA-D and joint KD~0.1/1/10/pure KD elsewhere. PCV uses shadow labels only; local has zero gain. All LCB decisions abstain. TCGA and MELD have only three test clients.}
\label{tab:policy-summary}
\setlength{\tabcolsep}{4pt}\begin{tabular}{llllr}
\toprule
Dataset & Four fixed gains (pp) & PCV choice & PCV gain [95\% CI] & Harm \\ \midrule
CREMA-D & -7.08 / -6.78 / -8.80 / -6.87 & local & $0.00$ [0.00,0.00] & 0.0\% \\
MELD & +11.16 / +16.05 / +17.66 / +17.89 & pure KD & $+17.89$ [15.30,22.03] & 0.0\% \\
UCF51-20 & +1.90 / +1.57 / -1.18 / -1.45 & local & $0.00$ [0.00,0.00] & 0.0\% \\
TCGA-BRCA & +13.33 / +17.62 / +17.18 / +14.21 & joint KD 0.1 & $+13.33$ [$-1.31$,30.00] & 33.3\% \\
UCI HAR & +3.23 / +3.88 / +1.98 / +1.69 & local & $0.00$ [0.00,0.00] & 0.0\% \\
\bottomrule\end{tabular}

\end{table}

\paragraph{Oracle and active-label controls.}
For accuracy, repeated random query halves alternate candidate selection and scoring. Balanced-accuracy splits are stratified; when a TCGA minority class has only one query example, a separate selection cohort chooses the policy and query labels remain evaluation-only. The exception is logged per cell. Active Testing uses the candidate-vote surrogate, expected zero-one-loss acquisition with a 0.1 uniform mixture, LURE correction, 200 acquisition repetitions and 5,000 client-cluster bootstrap replicates. TCGA uses class-conditional LURE ratio estimates and abstains until both classes are sampled. The main paper gives the label-budget curves and fixed-policy deployment table; Table~\ref{tab:policy-summary} retains the consolidated values alongside the detailed selection protocol.

\paragraph{Conditional evidence requirements.}
The simultaneous Hoeffding radius yields $m\geq\lceil2\log(K/\delta)/\gamma^2\rceil$ for radius at most mean gain $\gamma$. At $K=5,\delta=.05$, a gain of .20 requires 231 independent same-population shadow clients; three to fifteen clients cannot certify moderate gains. In a 50,000-trial bounded-gain diagnostic, deployment reaches 99.8\% at $m=384$ when the future mean is .20, yet becomes harmful in 99.8\% of trials when the future mean shifts down by .25. Larger cohorts address power under the population assumption; they cannot establish that assumption.

\section{Private Architectures and Compatible Prior Methods}
\label{sec:named-compatible}

\begin{table}[htbp]\centering\small
\caption{UCI HAR controls. Accuracy is \%; gains and intervals are pp. Panel A uses the sealed 12-subject study; Panel B uses compatible predefined models and each method's own reference. PeFLL receives 12 descriptor labels; ATP receives an unlabeled query batch. Panels do not share a baseline.}
\label{tab:har-controls}
\textbf{A. Selection and architecture sensitivity}\\[3pt]
\begin{tabular}{lrrrr}\toprule
Policy & Gain [95\% CI] & Harm & Linear & CNN \\ \midrule
fixed joint KD 1 & $+3.88$ [2.19,5.51] & 16.7\% & $-1.73$ & $+9.49$ \\
disagreement audit, 6 labels & $+3.22$ [1.83,4.64] & 8.3\% & $-1.08$ & $+7.52$ \\
disagreement audit, 24 labels & $+4.47$ [2.71,6.34] & 8.3\% & $-0.93$ & $+9.87$ \\
same-query oracle (diagnostic) & $+8.03$ -- & 0.0\% & $+2.89$ & $+13.18$ \\
\bottomrule\end{tabular}\par\medskip\textbf{B. Named compatible slices}\\[3pt]
\setlength{\tabcolsep}{4pt}\begin{tabular}{lllrrrr}
\toprule
Method & Role & Reference & Ref. acc. & Method acc. & Gain [95\% CI] & Harm \\ \midrule
PeFLL & shadow & FedAvg + fit (12 labels) & 45.9\% & 49.3\% & $+3.38\;[0.81,6.06]$ & 16.7\% \\
PeFLL & test & FedAvg + fit (12 labels) & 44.4\% & 51.2\% & $+6.79\;[3.35,9.95]$ & 16.7\% \\
ATP & shadow & Compatible global model & 42.48\% & 41.92\% & $-0.55\;[-1.39,0.00]$ & 83.3\% \\
ATP & test & Compatible global model & 46.02\% & 45.97\% & $-0.05\;[-0.69,0.59]$ & 50.0\% \\
\bottomrule\end{tabular}

\end{table}

The paired temporal-CNN-minus-linear interaction for fixed joint KD 1 is $+10.73\pp$ [3.21,17.73] on six shadow subjects (exact $p=.125$) and $+11.21\pp$ [7.57,15.17] on 12 test subjects ($p=.0005$). Each subject averages seeds before comparison. A retrospective architecture-specific shadow rule gains 3.06 points but harms 25.0\% of test subjects. Leave-one-subject-out selection over all 18 target subjects gains 4.30 points with 11.1\% harm when stratified, versus 3.06 points and 27.8\% when pooled. These analyses motivate a future prospective rule; they do not validate one on their hypothesis-discovery cohort.

\paragraph{PeFLL and ATP interfaces.}
PeFLL uses a learned example encoder to average 12 labeled examples into a descriptor, then generates all private parameters with a hypernetwork. Source inner updates provide first-order parameter-delta training signals. Separate modality/architecture banks restrict output to predefined linear/CNN spaces. All supervised controls receive the same labels, and development selects FedAvg-plus-fit as the reference. PeFLL gains 4.11/9.48 points on linear/CNN and 6.04/7.54 on fused/acceleration-only clients. ATP uses compatible FedAvg models, learns one adaptation rate per parameter tensor on source-selection clients, and restores the global state before one entropy step per unlabeled target batch of 20. Its architecture gains are $-.06/-.05$ points, and modality gains are $-.53/+.43$. This implementation omits custom differentiable BatchNorm-statistic rates because the HAR models do not expose them.

\paragraph{HeteroFedTTA reconstruction boundary.}
The linked anonymous implementation was unavailable when this study was built. We therefore reconstruct Algorithm 2 and Eq. 14. On the historical HAR slice, 12 source clients produce base/adapted responses on 754 source-query inputs, with labels excluded from the response loss. Its squared-mean adaptation loss approximates the native batch $\ell_2$ norm; the newer FedEMNIST reconstruction uses the native norm. At 300 steps, HAR shadow accuracy is 42.08\% for base-response KD, 41.39\% before dual-model adaptation, and 37.70\% afterward. Extending the compiler to 1,000 steps gives a best difference of $-1.34\pp$ [$-4.12$,1.41]; at 3,000 steps the compiled bases differ by only .09 points but target adaptation loses 5.53 points. This branch leaves test unopened. It is a bounded compatible-slice audit, not a claim about the published method's optimal performance.

The original FedEMNIST target-only TTA control selects entropy or fixed-pseudo-label updates, rates $10^{-5}$--$10^{-3}$ and 1/5/20 steps on 300 shadow writers. On the already evaluated test cohort, accuracy changes from 96.75\% to 96.65\%, $-0.10\pp$ [$-.26$,.05]. This historical post-hoc control uses a centralized source bank; the complete-pipeline study below is the primary nearest-method comparison.

\section{What Public Inputs Enable Compilation?}
\label{sec:support-details}

\paragraph{Shared resources and quantity.}
The 4,096 CelebA public images come from 3,137 identities outside all federation and evaluation clients, chosen by label-blind path-hash order. Anchor selection and response training do not use task labels. The historical teacher writes two class logits for each declared modality subset. Public-response validation uses its own disjoint inputs. The main figure displays the count frontier and controlled blur grid once; complete accuracy, gain, harm, agreement, coefficient and RMSE grids are in the released result files. At 4,096 anchors, teacher agreement is 85.37\%, coefficient norm-squared .074 and log-odds RMSE 3.307. The coefficient--gain association is $-.462$ pooled across budgets but $-.012$ [$-.095$,.070] after removing budget/backbone/modality means, and .001 [$-.075$,.075] at 4,096. It cannot reliably route individual clients.

\paragraph{Support interventions.}
The pose-restricted CelebA pool is the upper quartile of normalized nose displacement from the eye midpoint within the 3,584-image fit pool. Three seeds sample equal fit counts from broad and restricted pools. Neural models use 336 updates, batches of 32 images with two modality views, fixed late-block/head rates $10^{-4}/5\times10^{-4}$, FP32 cached prefixes and frozen BatchNorm. Response BCE on the common 512 validation images selects checkpoints every 56 updates. Table~\ref{tab:support-controls} joins response error, accuracy and harm in one view. EfficientNet at 128 images has uncertain accuracy change despite worse KL; all conditions and both signs remain visible.

\begin{table}[htbp]\centering\small
\caption{Equal-count support interventions. Accuracy is \%; differences and intervals are pp. Broad/restricted conditions share teacher, clients and training budget within each study. Intervals resample clients after averaging three draws. Restricted means pose-tail on CelebA and four scripts on CREMA-D. Harm columns are broad/restricted; KL difference is broad minus restricted; lower KL is better. Studies use different representations and local references.}
\label{tab:support-controls}
\setlength{\tabcolsep}{4pt}\begin{tabular}{lrrrrrr}
\toprule
Study/model & Fit & Broad acc. & Restr. acc. & Difference [95\% CI] & $\Delta$KL & Harm (\%) \\ \midrule
MobileNet blocks & 64 & 72.07 & 69.01 & 3.06 [1.84,4.32] & -0.0490 & 27.4/32.2 \\
MobileNet blocks & 128 & 78.46 & 75.42 & 3.04 [1.93,4.20] & -0.0426 & 17.8/24.8 \\
MobileNet blocks & 512 & 85.32 & 80.94 & 4.38 [3.26,5.54] & -0.0861 & 10.4/13.7 \\
EfficientNet blocks & 64 & 75.96 & 70.12 & 5.83 [4.51,7.21] & -0.0242 & 25.6/35.2 \\
EfficientNet blocks & 128 & 82.19 & 82.04 & 0.15 [-0.76,1.10] & +0.0248 & 14.1/14.4 \\
EfficientNet blocks & 512 & 87.82 & 84.94 & 2.89 [1.84,3.94] & -0.0667 & 8.5/13.0 \\
CelebA fixed ridge & 64 & 70.86 & 66.26 & 4.61 [3.67,5.56] & -0.0730 & 30.7/42.2 \\
CelebA fixed ridge & 128 & 75.05 & 71.16 & 3.90 [2.92,4.90] & -0.0604 & 22.2/30.0 \\
CREMA-D fixed features & 64 & 53.13 & 51.30 & 1.83 [-2.41,6.53] & -0.0001 & 100.0/100.0 \\
CREMA-D fixed features & 96 & 52.47 & 53.41 & -0.93 [-4.87,2.06] & +0.0044 & 100.0/100.0 \\
\bottomrule\end{tabular}

\end{table}

The short-step head-only control often selects its last checkpoint and is not used to establish representation superiority. After observing this, an explicitly post-hoc closed-form ridge control uses $\lambda=1$, identical FP32 initial features, the same fit anchors and both modality views. On broad support, late blocks exceed this control by 5.52/6.51/6.47 points at 64/128/512 images, with intervals [4.53,6.49]/[5.52,7.49]/[5.49,7.45]. The earlier fixed-ridge pose intervention has different feature handling and local references, so its rows are a separate study. Original head-only and ridge cells, including unfavorable conditions, remain in the numeric export.

For CREMA-D, six paired public actors are removed before training 54-client source banks. Five shadow actors select local settings and ten evaluate; only three evaluation actors have audio. The 478-clip public pool reserves 64 response-validation clips; four fixed scripts leave 114 eligible restricted-fit clips. A metadata-only amendment changes budgets from 64/128 to 64/96 before outcomes. Conditions share initialization, normalization and at most 720 updates. The loader materializes public annotations, but labels do not enter selection, loss or checkpointing. At 64 broad anchors, local/teacher/compiler accuracies are 80.68/55.19/53.13\%: the dominant 25.49-point deficit is already in supply. This is a small boundary case, not a second positive multimodal confirmation.

\paragraph{Alternative support and response controls.}
On 40 development identities, pivoted QR selects rank-diverse anchors but reaches 53.75\% versus hash-prefix 69.06\% at 64 inputs and 67.50\% versus 80.00\% at 256. Target-nearest selection reaches 63.91/77.97\% at those budgets; its .94-point advantage at 1,024 has an interval crossing zero. Soft prototypes reach 65.83\% at 4,096 validation anchors; on the ridge confirmation, hard-label ridge is 80.25\% versus response ridge 80.85\%. The evidence favors the public interface more than a unique soft-logit algorithm. A shadow-selected one-bit quantizer gives 84.91\% versus float32 85.28\%, then 85.71\% on a previously evaluated 300-identity cohort: $+22.56\pp$ over local, with 7.0\% harm. Its two-modality digest is 1,024 bytes (1-KB); this is post-hoc compression robustness, not privacy protection or a new prospective cohort.

\section{Complete-Pipeline Costs and Budget Sensitivity}
\label{sec:cost-details}

The comparison reuses three independently trained FedSGD source checkpoints, then forms an original bank, a meta-trained base bank, and a dual-response bank. The latter source stage persists client parameters across full writer sweeps. Each writer takes 20 public digest and 20 labeled meta-revisit steps; both response tables are updated after each sweep. A 32-writer pilot and 100 development writers select source inner rate .1 from .1/.5. Source budgets 1/2/4/8/16 and dual weights .2/.8/1 are chosen on development; the compiler retains inner rate .5. A plateau requires ordinary-KD and full-method development accuracy to vary by at most .25 points over the last three budgets, with at least four rounds. One seed reaches 16 rounds without meeting this criterion.

New compilers share initialization, optimizer, minibatch order and 3,584/512 public fit/validation images at 100/200 epochs. Full-method target rates .05/.1/.5, 1/5/20 steps and a no-update option are development-selected. Each adapted model retains the lowest-entropy checkpoint including step zero. Table~\ref{tab:pipeline-detail} reports per-architecture sensitivity and the source search. At 100/200 epochs, full-minus-minimal differences are $-.11$ [$-.28$,.07] and $.04$ [$-.12$,.21] points. At 200 epochs the meta-bank KD ablation gains .15 [.04,.27] over the original bank; adaptation does not inherit an automatic advantage from the bank improvement. Neither zero-crossing intervals nor an epoch cap establish equivalence or convergence.

\begin{table}[htbp]\centering\small
\caption{Complete-pipeline sensitivity and source effort. Panel A reports 100/200 epochs in each paired entry, retaining architecture-specific harm after seed averaging. Target time excludes I/O and loading. Panel B reports each source seed; the selected checkpoint and the full search are different costs. Times are sequential simulator measurements, not distributed latency.}
\label{tab:pipeline-detail}
\textbf{A. Private architecture and compiler budget}\\[3pt]
\setlength{\tabcolsep}{4pt}\begin{tabular}{llrrrr}
\toprule
Arch. & Pipeline & Accuracy (\%) & Harm (\%) & Compile (s) & Target (ms) \\ \midrule
MLP & Original + KD & 95.87/95.82 & 12.0/12.0 & 11.8/23.1 & 0.71/0.66 \\
MLP & Meta + KD & 95.74/95.97 & 12.7/11.0 & 11.0/22.4 & 0.66/0.66 \\
MLP & Meta + dual & 95.76/95.82 & 12.3/12.0 & 67.1/132.2 & 6.11/6.23 \\
CNN & Original + KD & 98.39/98.33 & 2.0/2.3 & 12.9/25.7 & 0.79/0.77 \\
CNN & Meta + KD & 98.44/98.49 & 2.3/2.7 & 12.8/25.3 & 0.69/0.69 \\
CNN & Meta + dual & 98.29/98.42 & 2.7/2.0 & 67.9/136.5 & 42.88/10.94 \\
\bottomrule\end{tabular}

\par\medskip\textbf{B. Source-bank construction and search}\\[3pt]
\begin{tabular}{rrrrrr}\toprule Seed & source rounds & FedSGD (s) & extra source (s) & budget search (s) & plateau \\ \midrule

81011 & 2 & 91.6 & 5349.6 & 12463.9 & yes \\

81012 & 4 & 87.3 & 10668.1 & 23589.8 & yes \\

81013 & 1 & 85.7 & 2656.9 & 43954.6 & no \\

\bottomrule\end{tabular}
\end{table}

The common FedSGD stage uses 614,400 labeled presentations. Each additional source round uses 1,280,000 labeled revisit presentations, 1,280,000 public digest presentations and 4,096,000 ensemble evaluations. Search time includes compiler trials and development scoring, excluding final validation/test scoring. Original-bank export additionally takes .38/.10/.08 seconds across seeds; regenerated responses exactly match the stored tables. Source computation differs and is charged explicitly. Compilation is reusable for a chosen architecture; adaptation recurs per newcomer. Per-seed harm is retained in the CSV/JSON export, including MLP original-KD rates 12.3/13.7/14.0\% at 100 epochs.

FedEMNIST public pixels occupy 3,211,264 bytes; each ten-class float32 response table occupies 163,840 bytes (160 KiB). Uncached original/full handoffs are 3,375,104/3,538,944 bytes, a 4.85\% difference, even though response-only payload doubles. CelebA's JPEG pool occupies 28.2 MB and its two binary-response tables total 64 KiB. No distributed network latency was measured. Onboarding timings exclude data preparation and loading as declared; sequential simulator timings do not measure distributed deployment speed.

\section{Where Does Client Harm Come From?}
\label{sec:harm-details}

For a fixed teacher, $S_c=U(h_c)-U(f_{c,0})$, $C_c=U(h_c)-U(f_c)$ and $G_c=S_c-C_c$. Inherited harm is $S_c<0,G_c<0$; compiler-induced harm is $S_c\geq0,G_c<0$. These events partition observed harm. Table~\ref{tab:benefit-safety} refers to the historical local references in Table~\ref{tab:cohorts}; it must not be mixed with the strengthened or different-budget endpoints in main Table~2.

\begin{table}[htbp]\centering\small
\caption{Historical harm decomposition using each cohort's original local reference. Supply, loss and gain are pp; harm is a percentage after architecture averaging. All quantities use query labels and are diagnostic.}
\label{tab:benefit-safety}
\begin{tabular}{lrrrrrr}\toprule Study & Supply & Compile loss & Gain & Inherited & Compiler & Total \\ \midrule
CelebA ridge & $+23.90$ & $7.27$ & $+16.63$ & 4.0\% & 11.3\% & 15.3\% \\
CelebA blocks & $+25.02$ & $1.71$ & $+23.31$ & 4.0\% & 2.0\% & 6.0\% \\
CelebA decorrelated & $+10.78$ & $1.12$ & $+9.66$ & 10.3\% & 9.3\% & 19.7\% \\
FedEMNIST & $+21.44$ & $2.59$ & $+18.85$ & 0.3\% & 1.0\% & 1.3\% \\
\bottomrule\end{tabular}
\end{table}

For the decorrelated cohort, the original strict floating-point sign gave 20.7\% harm. Treating absolute differences below $10^{-12}$ as ties gives 59/300 (19.7\%), split into 31 inherited and 28 compiler-induced cases. The original raw result remains unchanged. On the same binary block cohort, frozen-feature versus updated-block compilation losses are 6.90/1.71 points and compiler-induced harm is 6.3/2.0\%. Missing landmarks do not drive this improvement: image-only and image--landmark gains are 24.74/22.03 points, with compiler-induced harm 2.1/1.9\%. These paired results must not be compared as interventions against the separate ridge cohort.

On the original FedEMNIST model, per-architecture harm is 6.0\% for MLP and 1.0\% for CNN, while averaging architectures before checking sign gives 1.3\%. All 234 writers with at least 95\% teacher agreement have nonnegative observed gain, but this threshold was discovered post hoc. Exact response recovery would set compilation loss to zero while leaving teacher benefit unidentified. None of these observations certifies every future client's risk.

\section{Visual Comparisons of Reported Results}
\begin{figure*}[t]
\centering
\includegraphics[width=\textwidth]{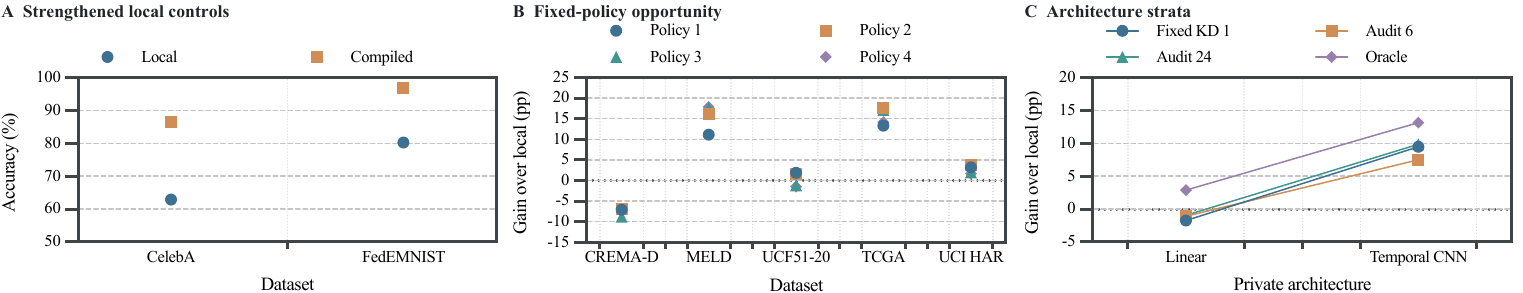}
\caption{\textbf{Utility, candidate opportunity and private architecture.}
\textbf{A:} Strengthened local controls versus compilation on separate CelebA and FedEMNIST cohorts.
\textbf{B:} Four fixed policies on each dataset. Policies 1--4 denote global, cluster 0, cluster 1 and uniform on CREMA-D; elsewhere they denote KD weights 0.1, 1, 10 and pure KD. These are dataset-specific candidate menus.
\textbf{C:} UCI HAR gains by private architecture under fixed KD, 6- and 24-label disagreement audits, and a same-query diagnostic oracle. The oracle is not an allowed deployment method. Each comparison retains its original local reference.}
\label{fig:prism-comparisons}
\end{figure*}

\begin{figure*}[t]
\centering
\includegraphics[width=.90\textwidth]{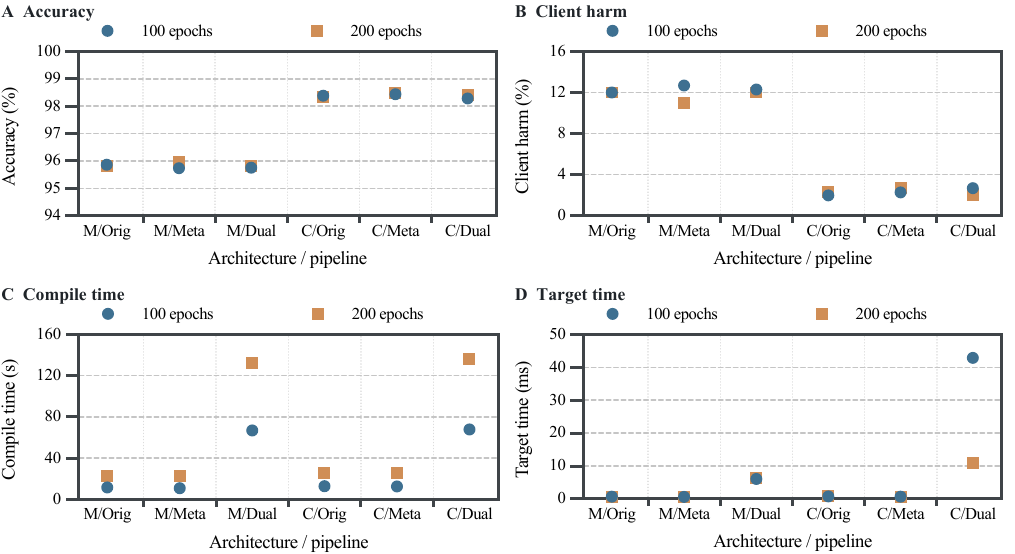}
\caption{\textbf{Complete-pipeline budget sensitivity.} Accuracy, client harm, compilation time and target inference time at 100 and 200 compilation epochs. M/C denote pixel-MLP/CNN; Orig, Meta and Dual denote original+KD, meta+KD and meta+dual pipelines. Points are the reported three-seed means. Timing is sequential simulator time and excludes loading and I/O; it is not a parallel deployment latency estimate.}
\label{fig:prism-budget}
\end{figure*}

\begin{figure*}[t]
\centering
\includegraphics[width=\textwidth]{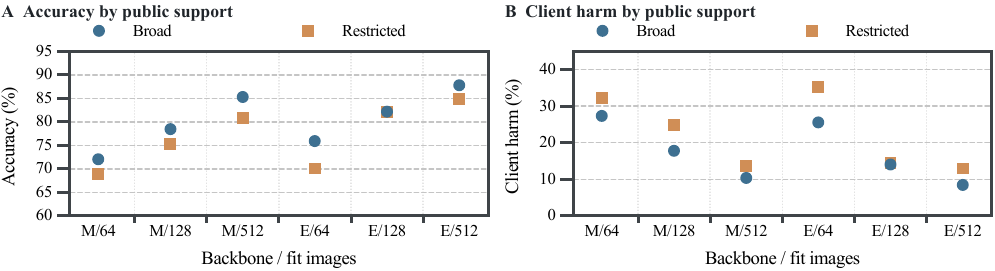}
\caption{\textbf{Equal-count neural public-support comparison.} Broad versus pose-restricted public inputs at 64, 128 and 512 fit images. M/E denote MobileNetV3-Small/EfficientNet-B0. Each setting uses 336 neural updates, 512 additional shared validation responses, three anchor draws and the same 270 shadow identities. Panels show accuracy and architecture-specific client harm; the main support figure reports paired accuracy differences with client-bootstrap intervals.}
\label{fig:prism-support}
\end{figure*}

Figures~\ref{fig:prism-comparisons}--\ref{fig:prism-support} visualize results already tabulated in the main paper and this supplement. They add no experimental runs. Points use the published rounded means; only the paired differences in the main support figure carry confidence intervals.

\section{Identifiability Proofs and Their Scope}
\label{sec:anchor-details}

For completeness, let $Z_A\in\mathbb R^{n\times d}$ contain the augmented private anchor features as rows.  Under linear realizability, $t(x)=w_*^\top z(x)$.  If $z(x)=Z_A^\top\alpha_x$ for every target input, then any $\hat w$ satisfying $Z_A\hat w=Z_Aw_*$ also satisfies
\begin{equation}
  z(x)^\top\hat w
  =\alpha_x^\top Z_A\hat w
  =\alpha_x^\top Z_Aw_*
  =t(x).
\end{equation}
Conversely, if a target $z(x_0)$ is outside the row span of $Z_A$, the fundamental theorem of linear algebra gives a $v\in\operatorname{null}(Z_A)$ for which $v^\top z(x_0)\neq0$.  The two heads $w_*+v$ and $w_*-v$ produce identical anchor responses and different target scores, so no response-only compiler can identify which is correct.  This establishes necessity and sufficiency.

With imperfect fitting and $z(x)=Z_A^\top\alpha_x$, Cauchy--Schwarz gives
\begin{equation}
  |z(x)^\top(\hat w-w_*)|^2
  \leq \|\alpha_x\|_2^2\,\|Z_A(\hat w-w_*)\|_2^2.
\end{equation}
For the empirical ridge diagnostic we use the regularized coefficient
\begin{equation}
  \alpha_x=Z_A(Z_A^\top Z_A+\lambda I)^{-1}z(x),
\end{equation}
For $\lambda>0$, this coefficient generally does not satisfy $Z_A^\top\alpha_x=z(x)$. Defining $r_\lambda(x)=z(x)-Z_A^\top\alpha_x$ and $\Delta w=\hat w-w_*$ gives the valid bound
\begin{equation}
 |z(x)^\top\Delta w|\leq
 \|\alpha_x\|_2\|Z_A\Delta w\|_2+
 \|r_\lambda(x)\|_2\|\Delta w\|_2.
\end{equation}
We report $\mathbb E\|\alpha_X\|_2^2$ only as a regularized geometric diagnostic, not an exact span certificate. The exact bound above uses a coefficient satisfying the representation equality (e.g., a pseudoinverse solution when feasible). Neither diagnostic implies label accuracy under misspecification.

For $C$ classes, apply the same statement to each logit contrast against one fixed reference class.  The required span condition remains a property of the newcomer features; the response target becomes a $C-1$ column matrix.  The binary experiment uses the single class-1 versus class-0 log-odds contrast.

\paragraph{Unlabeled risk-order indistinguishability.}
Fix $P_X$ and all candidate classifiers. In world $P$, set $Y=f_0(X)$ everywhere, making the local classifier error-free and every genuinely disagreeing candidate no better. In world $Q$, set $Y=f_j(X)$ on $D_j=\{x:f_j(x)\ne f_0(x)\}$ and keep $Y=f_0(X)$ elsewhere. The marginal inputs and all observable predictions are identical in the two worlds. If $P_X(D_j)>0$, selecting $f_j$ has strictly negative gain in $P$ and strictly positive gain in $Q$. Since the selector sees identical evidence, it cannot guarantee safe nontrivial transfer over unrestricted label conditionals. This argument requires disagreement on positive target mass, not merely different parameter values. The anchor-span proof and regularized residual are given above.

\paragraph{Conditional mean-gain concentration.}
Condition on the frozen bank and policies $\mathcal F$. For independently drawn same-stratum shadow clients and fixed candidate $k$, gains lie in $[-1,1]$, so Hoeffding gives
\[
 \Pr\!\left(\bar G_k-\mathbb E[G_k(C)\mid\mathcal F]\geq\epsilon\mid\mathcal F\right)
 \leq \exp(-m\epsilon^2/2).
\]
A union bound over $K$ fixed candidates and $\epsilon=\sqrt{2\log(K/\delta)/m}$ yields the stated simultaneous bound. Candidate or bank selection using the same shadow labels invalidates this fixed-policy argument unless independently accounted for. The guarantee concerns population mean gain under the declared distribution, not each client or an arbitrarily shifted future population.

\section{Related Interfaces and Reproduction Entry Points}

Table~\ref{tab:applicability} records task-level applicability rather than
assigning artificial failure scores to methods whose required interface is
absent.  A check means that the cited work explicitly treats the capability;
parentheses mark a restricted form.  ``Frozen'' means that onboarding does not
update the completed federation.  ``Private output'' means that improvement is
realized in a newcomer-side model rather than only a server/global model.

\begin{table}[h]
\centering
\small
\caption{Novelty and applicability boundary.}
\label{tab:applicability}
\begin{tabular}{lccccc}
\toprule
Method & unseen & frozen & multimodal & unseen arch. & private output \\
\midrule
OD-PFL \citep{amosy2024odpfl} & $\checkmark$ & $\checkmark$ & -- & -- & (predefined) \\
UPFL/HeteroFedTTA \citep{ye2024upfl} & $\checkmark$ & $\checkmark$ & -- & $\checkmark$ & $\checkmark$ \\
FLowDUP \citep{zakerinia2026flowdup} & $\checkmark$ & $\checkmark$ & -- & -- & (predefined) \\
EPFL \citep{liu2025epfl} & $\checkmark$ & $\checkmark$ & -- & -- & (predefined) \\
PeFLL \citep{scott2024pefll} & $\checkmark$ & $\checkmark$ & -- & -- & (predefined) \\
pFedHN \citep{shamsian2021pfedhn} & $\checkmark$ & $\checkmark$ & -- & -- & (predefined) \\
HyperFLoRA \citep{lu2024hyperfloRA} & $\checkmark$ & $\checkmark$ & -- & -- & (adapter) \\
HyperFedZero \citep{zhou2026hyperfedzero} & $\checkmark$ & $\checkmark$ & -- & -- & (predefined) \\
ATP \citep{bao2023atp} & $\checkmark$ & $\checkmark$ & -- & -- & (global copy) \\
TFL \citep{ma2024tfl} & $\checkmark$ & $\checkmark$ & -- & -- & -- \\
TGFed \citep{niu2026tgfed} & $\checkmark$ & $\checkmark$ & -- & -- & (compatible) \\
COTE \citep{zhu2026cote} & $\checkmark$ & $\checkmark$ & (V--L) & -- & (prompt) \\
DP$^2$FL \citep{chang2025dp2fl} & $\checkmark$ & (inference) & (V--L) & -- & (prompt) \\
HAFL-GHN \citep{litany2022ghn} & $\checkmark$ & $\checkmark$ & -- & (graph family) & $\checkmark$ \\
FedMBridge \citep{chen2024fedmbridge} & -- & -- & $\checkmark$ & (training clients) & $\checkmark$ \\
FedAFD \citep{tan2026fedafd} & -- & -- & $\checkmark$ & (training clients) & $\checkmark$ \\
FedMD \citep{li2019fedmd} & -- & -- & -- & $\checkmark$ & $\checkmark$ \\
pFedDSH \citep{nguyen2026onboarding} & $\checkmark$ & -- & -- & -- & (hypernetwork) \\
\method{} task & $\checkmark$ & $\checkmark$ & $\checkmark$ & $\checkmark$ & $\checkmark$ \\
\bottomrule
\end{tabular}
\end{table}

PeFLL and ATP are executed on their compatible interfaces in Section~\ref{sec:named-compatible}; HeteroFedTTA is evaluated through explicitly scoped reconstructions. Response distillation itself is established prior work. The contribution is the post-federation benchmark, its information-access contract and controlled evidence about benefit and portability. The strongest confirmations address visual architecture portability; image-derived landmarks do not establish independent-sensor multimodal success.

\paragraph{Reproduction and full numeric records.}
Run \texttt{python3 fednbo.py verify} for registered hashes, exact CelebA cohort identities and role separation; \texttt{verify --data} checks local assets. \texttt{run ... --dry-run} prints registered training commands, and \texttt{results} locates current summaries. The README maps cohorts and allowed information. Main and supplementary tables are generated from recorded evidence; \texttt{paper\_cvpr/compose\_evidence.py --check} verifies the compact displays. The directory \path{paper_cvpr/evidence_tables/} contains machine-readable full condition grids, contrasts and the previous table archive. Existing per-client JSONL records retain individual seeds, interfaces and unfavorable results.

The primary evidence resides in \path{results/revision_20261007/}; fixed-feature support and source-budget audits in \path{results/acceptance_revision_20261008/}; complete-pipeline and neural-support records in \path{results/completion_revision_20261009/}. The registry binds the original policy studies, confirmation cohorts and compatible baselines. These grouped entry points replace a repeated list of every intermediate file. The original compiler source was recovered from its creation patch and matches its frozen hash; later quantization extensions are kept separately. Full neural reproduction requires separately prepared public datasets and weights. The evidence bundle verifies records and code without redistributing raw data or checkpoints.

\paragraph{Ethics and data governance.}
The study uses public benchmark data and previously extracted features. TCGA site identifiers define experimental clients and do not assess institutions. IDC/ILC classification is a methodological probe, not a clinical diagnostic system. Response release is assumed permitted; differential privacy and leakage protection are outside the evaluated contract.

{\small\bibliographystyle{ieeenat_fullname}\bibliography{references}}